\documentclass[10pt,twocolumn,letterpaper]{article}

\usepackage{cvpr}              
\usepackage{microtype}
\usepackage{multirow}
\definecolor{cvprblue}{rgb}{0.21,0.49,0.74}
\usepackage[pagebackref,breaklinks,colorlinks,allcolors=cvprblue]{hyperref}

\def\paperID{11555} 
\def\confName{CVPR}
\def\confYear{2025}

\title{Decoder Gradient Shield: Provable and High-Fidelity Prevention of Gradient-Based Box-Free Watermark Removal}

\author{
Haonan An$^1$ \quad Guang Hua$^{2}$\thanks{Equal Contribution} \quad Zhengru Fang$^1$ \quad Guowen Xu$^3$ \\
Susanto Rahardja$^2$ \quad Yuguang Fang$^1$ \\
$^1$Department of Computer Science, City University of Hong Kong, Hong Kong \\
$^2$Infocomm Technology (ICT) Cluster, Singapore Institute of Technology (SIT), Singapore 138683 \\
$^3$School of Computer Science and Engineering, University of Electronic Science and Technology of China \\
}

\begin{document}
\maketitle

\begin{abstract}
The intellectual property of deep image-to-image models can be protected by the so-called box-free watermarking. It uses an encoder and a decoder, respectively, to embed into and extract from the model's output images invisible copyright marks. Prior works have improved watermark robustness, focusing on the design of better watermark encoders. In this paper, we reveal an overlooked vulnerability of the unprotected watermark decoder which is jointly trained with the encoder and can be exploited to train a watermark removal network. To defend against such an attack, we propose the decoder gradient shield (DGS) as a protection layer in the decoder API to prevent gradient-based watermark removal with a closed-form solution. The fundamental idea is inspired by the classical adversarial attack, but is utilized for the first time as a defensive mechanism in the box-free model watermarking. We then demonstrate that DGS can reorient and rescale the gradient directions of watermarked queries and stop the watermark remover's training loss from converging to the level without DGS, while retaining decoder output image quality. Experimental results verify the effectiveness of proposed method. Code of paper will be made available upon acceptance.
\end{abstract}

\section{Introduction}
\label{sec:intro}

Today's deep learning models can provide exceptional performance across a wide range of tasks, even surpassing human capability \cite{fang2024pacp, zhang2024smartcooper, Hou2023Federated, an2024channel, fang2025ton}. However, these resource-intensive models are also subject to the risk of intellectual property infringement. To address this problem, model watermarking has been developed to verify model ownership or detect model theft.

According to how the watermark is extracted, model watermarking can be classified into white-box, black-box, and box-free methods. White-box methods \cite{uchida2017embedding,wang2022rethinking,Lv2023A} require access to the protected model's internal content in which the watermark is encoded. Black-box methods \cite{adi2018turning,Li2023Universal,hua23unambiguous}, also known as backdoor watermarking, require querying the protected model for watermark extraction since they encode the watermark into the model's input-output mapping. Box-free methods, however, extract the watermark directly from the protected model's outputs, which are more flexible and are specially suitable for models that generate high-entropic content, e.g., encoder-decoder image models \cite{wu2020watermarking,zhang2022deep,zhang2024robust} and generative adversarial networks (GANs) \cite{Lin2024A,Fei2024Wide}. Since only box-free methods embed watermarks into generated outputs, they present a viable solution to the growing demand for AI-generated content attribution.

\begin{figure}[!t]
    \centering
    \includegraphics[width=1\linewidth]{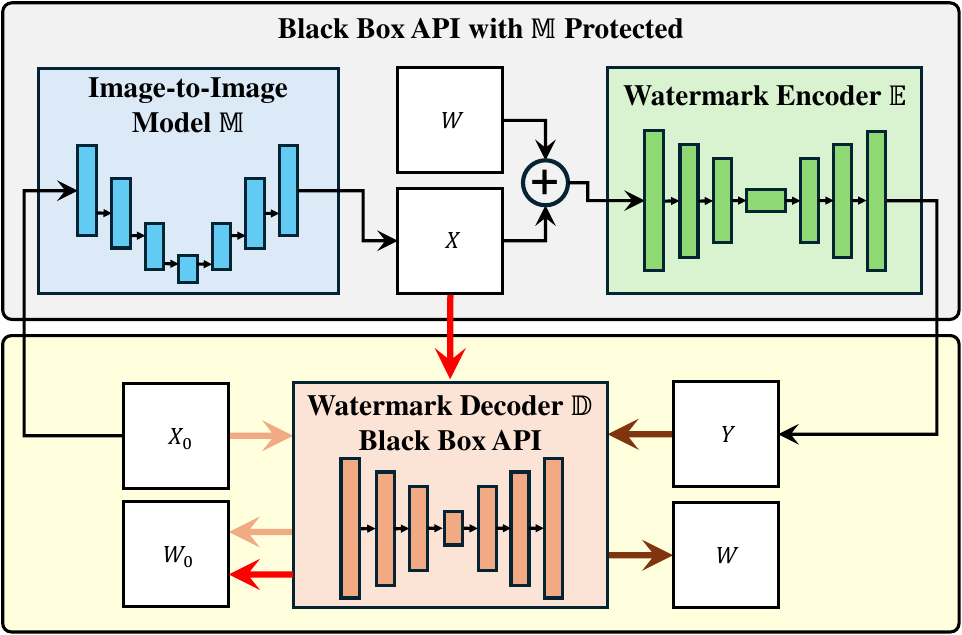}
    \caption{Flowchart of box-free model watermarking for image-to-image models. The thin black arrows represent the black-box querying flow (processing and watermarking), while the thick colored arrows represent potential watermark extraction, and each colored arrow pair corresponds to a single input-output pair for $\mathbb{D}$.}
    \label{fig:victim_architecture}
\end{figure}

The typical workflow of box-free watermarking for image-to-image models is depicted in Figure \ref{fig:victim_architecture} \cite{zhang2022deep}. Considering a model $\mathbb{M}$ that takes an input $X_0 \in \mathcal{X}_0$ and generates an output $X \in \mathcal{X}$, box-free watermarking creates a watermark encoder $\mathbb{E}$ which embeds a copyright mark image $W$ into $X$ and yields the watermarked image $Y\in\mathcal{Y}$. A dedicated watermark decoder $\mathbb{D}$ is jointly created and can extract the mark $W$ from the watermarked set $\mathcal{Y}$ or a null-mark $W_0$ from the non-watermarked complement set $\mathcal{Y}^\complement$. According to the specific image task, $(\mathcal{X}_0, \mathcal{X})$ can be (noisy, denoised), (original style, transferred style), etc. In the above process, the model $\mathbb{M}$ is protected by only providing $Y$, instead of $X$, to the user. It has been verified that if the attacker uses the collection of $X_0$ and $Y$ to train a surrogate model, $W$ can still be extracted by $\mathbb{D}$ from the surrogate output images \cite{zhang2022deep}. We refer to $\mathbb{M}$ as either the protected model or the victim model where applicable.

\begin{figure}[!t]
    \centering
    \includegraphics[width=1\linewidth]{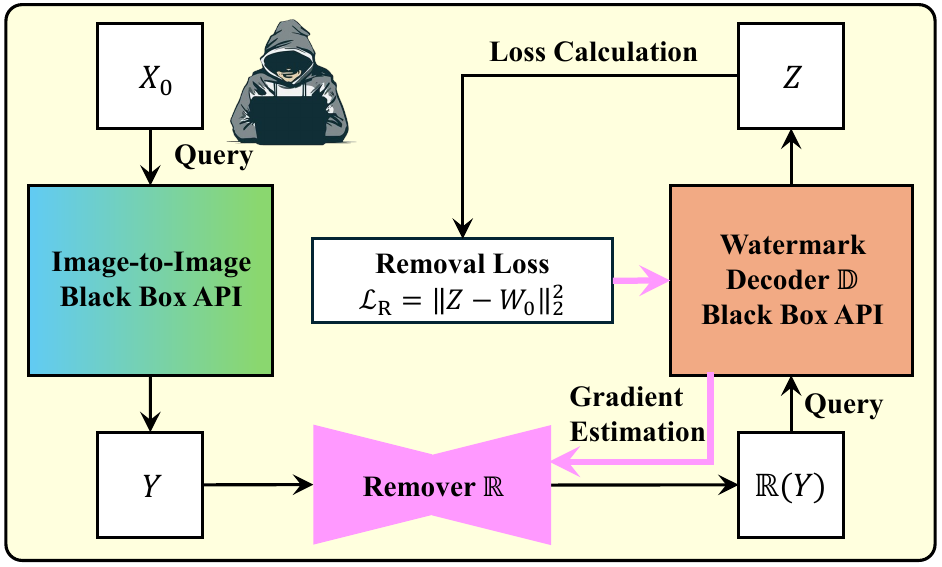}
    \caption{Flowchart of gradient-based removal attack. The gradient backpropagated from $\mathbb{D}$ can be estimated by leveraging black-box adversarial attacks. In our setting, however, we assume the attacker can directly obtain the gradient without estimation.}
    \label{fig:gradient_based_attack}
\end{figure}

The attacker has the freedom to alter $Y$ with the hope of removing the watermark while preserving image quality, prior to surrogate training. Intuitively, the alteration can be compression, noise addition, flipping, cropping, etc. Although this vulnerability can be mitigated by adding an augmentation layer between $Y$ and $\mathbb{D}$ when training $\mathbb{E}$ and $\mathbb{D}$ \cite{zhang2024robust}, the attacker can launch a more advanced removal attack by training a removal network $\mathbb{R}$, as shown in Figure \ref{fig:gradient_based_attack}. Adversarial attack literature \cite{Ilyas2018Black, Dong2022Query} has shown that the gradient of $\mathbb{D}$ can be estimated via black-box queries, and the gradient can then be used to train $\mathbb{R}$. Such a gradient-based attack is feasible because $\mathbb{D}$ is jointly trained with $\mathbb{E}$ and contains the watermarking mechanism that can be compensated.

In this paper, we first show that the above gradient-based attack can remove state-of-the-art box-free watermarks. Then, under the practical threat model with the gradient information of $\mathbb{D}$ assumed to be observable to the attacker, we propose a more advanced defense mechanism called decoder gradient shield (DGS). For non-watermarked queries, the black-box API of $\mathbb{D}$ returns the null watermark output (close to the all-white image $W_0$) to the users. For watermarked queries, DGS reorients and rescales the corresponding gradient so that when the gradient is backpropagated and used to train $\mathbb{R}$, the training loss will not be able to converge to the level without DGS protection. Such reorientation is realized by adding specially crafted perturbations on the output of $\mathbb{D}$ while retaining output image quality. Notably, our approach could yield a neat closed-form solution, which is distinct from existing defense or related solutions. Our contributions are summarized as follows.

\begin{itemize}
\item We reveal the vulnerability of the unprotected watermark decoder in existing box-free watermarking methods, that is, the decoder can be exploited via black-box access to train a watermark removal network to remove existing box-free watermarks.

\item We propose a novel DGS framework to \emph{prevent} model extraction from exploiting the watermark decoder. This is different from the existing \emph{post-hoc} methods that extract watermarks from successful surrogates.

\item We provide both the closed-form solution and extensive experimental results to verify and demonstrate the effectiveness of our proposed method.
\end{itemize}

\section{Related Work}
\label{sec:related work}
\subsection{Model Extraction}
Model extraction, also known as model stealing or surrogate attack, aims to replicate a victim model's functionality, in which the attacker curates a set of query data, and with the victim model returned query outputs, trains a surrogate model. Previous research mainly focused on stealing deep classification models, in which the surrogate query data can be public data \cite{Orekondy2019Knockoff}, evolutionary proxy data \cite{Barbalau2020Black}, or even synthetic noise \cite{Truong2021Data,Kariyappa2021Maze}, and the attack can be successful even if the victim model only returns hard labels \cite{Sanyal2022Towards}. Model extraction can also be launched against self-supervised image encoder \cite{Sha2023Cant} and ensemble models \cite{Ma2023DivTheft}.

\subsection{Box-Free Watermarking}
Box-free watermarking is so named because the watermark is extracted from the model outputs using a dedicated watermark decoder, not requiring the protected model. The watermark decoder can be (1) pretrained and frozen when fine-tuning the protected model for watermark embedding \cite{Lin2024A,Fei2024Wide}, (2) jointly trained with the protected model \cite{wu2020watermarking,Lukas2023PTW}, or (3) a post-hoc model not coupled with the protected model \cite{zhang2022deep,zhang2024robust}. Since the watermarks are embedded in the protected model outputs, box-free methods are commonly applied to generative models with high entropic image outputs.

\subsection{Watermark Removal}
Watermark removal in the context of box-free watermarking is similar yet different from the conventional image watermarking. Through the alteration of victim model returned outputs before surrogate training, the removal attack aims to ensure that the watermark cannot be extracted from surrogate model generated images. Such alteration can be either normal image augmentation \cite{zhang2022deep} or a specially designed process such as inpainting \cite{liu2023erase} or an image-to-image watermark remover. These removers are also constrained to preserve the image quality for effective surrogate training.

\subsection{Gradient Attack and Defense}
The attack on neural network gradient has been researched for a decade and is mainly for generating adversarial examples. It can be a white-box attack such as the classic fast gradient sign method (FGSM) \cite{Goodfellow2014Explaining} and projected gradient descent (PGD) \cite{madry2018towards}. Under the more practical black-box setting, the gradient to be attacked can be estimated via querying \cite{Ilyas2018Black}, while the required query times can be substantially reduced \cite{Dong2022Query}. These results serve as the foundation of our threat model assuming observable gradients of the black-box decoder to the attacker. In model watermarking, gradient alteration has instead been utilized as a defense approach to protect the extraction of classification models \cite{Mantas2022How,pp_2020_iclr_defense_model_stealing}, but it cannot withstand hard-label based extraction \cite{Sanyal2022Towards}. We note that our work is the first to incorporate gradient alteration in protecting image-to-image models.

\section{Problem Formulation}
\label{sec:problem_formulation}
\subsection{Box-free Model Watermarking}
\label{sec:basic}
Since all box-free watermarking methods share the same extraction process using $\mathbb{D}$, without loss of generality, we consider the post-hoc type \cite{zhang2022deep,zhang2024robust} as our watermarking model, whose workflow is depicted in Figure \ref{fig:victim_architecture}. It contains

\begin{itemize}
\item Image Processing: $X \triangleq \mathbb{M}(X_0)$,

\item Watermark Embedding: $Y \triangleq \mathbb{E}({\rm{Concat}}(X, W))$,

\item Watermark Extraction: $\mathbb{D}: \mathcal{Y} \to W$ and $\mathcal{Y}^\complement \to W_0$,
\end{itemize}
where ${\rm{Concat}}(\cdot)$ denotes the channel-wise concatenation and $\mathcal{X}_0,\mathcal{X} \subset \mathcal{Y}^\complement$. With $\mathbb{M}$ untouched, $\mathbb{E}$ and $\mathbb{D}$ are jointly trained by minimizing
\begin{equation}
    \label{eq:loss_joint}
    \mathcal{L}_{\text{Victim}} = \alpha_1 \mathcal{L}_{\text{Embed}} + \alpha_2 \mathcal{L}_{\text{Fidelity}},
\end{equation}
where $\alpha_1$ and $\alpha_2$ are weighting parameters,
\begin{align}
    \mathcal{L}_{\text{Embed}} & = \sum_{Y \in \mathcal{Y}}\|\mathbb{D} \left(Y \right) - W \|_2^2 +  \sum_{S \in \mathcal{Y}^\complement}\|\mathbb{D} \left(S \right) - W_0 \|_2^2, \label{eq:loss_robust}\\
    \mathcal{L}_{\text{Fidelity}} & = \sum_{Y \in \mathcal{Y}, X \in \mathcal{X}} \|Y - X \|_2^2,  \label{eq:loss_fidelity}
\end{align}
and $S$ denotes an arbitrary image. Note that $\mathbb{E}$ is indirectly expressed by its output $Y$ in (\ref{eq:loss_robust}) and (\ref{eq:loss_fidelity}), which facilitates our subsequent attack formulation in which $\mathbb{E}$ is inaccessible.

\subsection{Threat Model}
\textbf{Defender}. We consider the owner of $\mathbb{M}$ as the defender, who not only trains $\mathbb{M}$ but also implements box-free watermarking and owns $\mathbb{E}$ and $\mathbb{D}$. The defender aims to extract watermarks in surrogate model generated images or prevent surrogate training, i.e., incurring non-negligible performance degradation in surrogate models. As shown in Figure \ref{fig:victim_architecture}, the defender only provides the black-box API which accepts query $X_0$ and only returns $Y$, while $\mathbb{M}$ and $\mathbb{E}$ are strictly private. The black-box API of $\mathbb{D}$ is also provided to accept watermark verification queries. 

\textbf{Attacker.} On the other side, the attacker aims to extract $\mathbb{M}$ with a watermark-removed surrogate model. To achieve so, the attacker curates a set of $X_0$ to query the victim model and obtains a set of $Y$, which is identical to how normal users behave. Then, prior to surrogate training, the attacker alters $Y$ for watermark removal while preserving image quality. Meanwhile, the attacker can query $\mathbb{D}$ to check if an image contains the defender-embedded watermark. In addition, we assume that the attacker can observe the gradient backpropagated from the output of $\mathbb{D}$, thanks to the adversarial attack literature. 
Furthermore, if the watermark remains intact despite attempts by a removal-loss-minimized remover, the attacker might reasonably infer the presence of defender-imposed gradient perturbations and adjust the returned gradient as a countermeasure.


\subsection{Gradient-based Removal Attack}
We now formulate the gradient-based box-free watermark removal attack depicted in Figure \ref{fig:gradient_based_attack}. The rationale behind is that the attacker can deploy an inverse of $\mathbb{E}$, i.e., another image-to-image network $\mathbb{R}$, which takes in $Y$ and undoes watermark embedding. This can be achieved by using $\mathbb{D}$ as a watermark verifier and an all-white null watermark $W_0$ as the supervision signal, leading to the minimization problem with the loss function
\begin{equation}
\label{eq:loss_attacker}
\mathcal{L}_\text{Attack} = \beta_1 \mathcal{L}_\text{Removal} + \beta_2 \mathcal{L}_\text{Fidelity}^{\text{A}},
\end{equation}
where $\beta_1$ and $\beta_2$ are the weighting parameters,
\begin{align}
\mathcal{L}_\text{Removal} & =  \|\mathbb{D}[\mathbb{R}(Y)] - W_0\|_2^2, \label{eq:r_loss}\\
\mathcal{L}_\text{Fidelity}^{\text{A}} & = \|\mathbb{R}(Y) - Y\|_2^2 \label{eq:a_fidelity_loss}. 
\end{align}
The removal loss ensures that the altered $Y$, i.e., $\mathbb{R}(Y)$, does not contain the watermark, while the fidelity loss ensures the preserved image quality for subsequent surrogate training. To minimize (\ref{eq:r_loss}), it holds that 
\begin{equation}\label{eq:attack_gradient}
\nabla {\mathcal{L}_{{\text{Removal}}}} \propto \frac{{\partial {\mathcal{L}_{{\text{Removal}}}}}}{{\partial \mathbb{D}\left[ {\mathbb{R}(Y)} \right]}}\frac{{\partial \mathbb{D}\left[ {\mathbb{R}(Y)} \right]}}{{\partial \mathbb{R}(Y)}},
\end{equation}
where $\nabla {\mathcal{L}_{{\text{Removal}}}}$ is the gradient component backpropagated to update $\mathbb{R}$ parameters. The $\ell_2$-norm loss function is used in (\ref{eq:r_loss}) in this paper, although other loss functions can also be possible. Under our threat model, the gradient of $\mathbb{D}$ in  (\ref{eq:attack_gradient}), ${\partial \mathbb{D}\left[ {\mathbb{R}(Y)} \right]}/{\partial \mathbb{R}(Y)}$, is observable, and all other gradient components are determined (known) by the attacker. Therefore, $\mathbb{R}$ can be effectively trained in the black-box setting. 

We note that the above attack resembles the classical adaptive filter for model inversion \cite{Haykin2002Adaptive} where $\mathbb{R}$ is the inverse of $\mathbb{E}$. Given the black-box access of $\mathbb{D}$ and without protection, this attack has a theoretical guarantee of convergence.

\section{The Proposed Decoder Gradient Shield}
\begin{figure*}[!t]
    \centering
    \includegraphics[width=.98\linewidth]{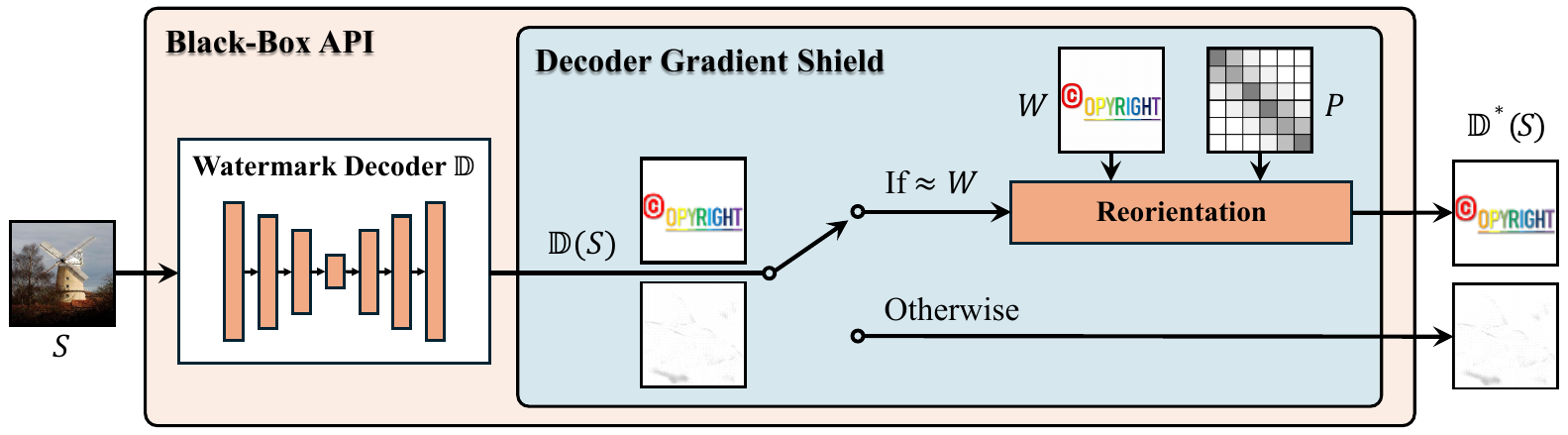}
    \caption{Flowchart of the proposed DGS in the black-box API of $\mathbb{D}$.}
    \label{fig:gsp_architecture}
\end{figure*}
\label{sec:proposed_defense}
The black-box setting of $\mathbb{D}$ enables the defender to alter its output before returning to the user for protection purposes, while the alteration is subject to the constraint of not affecting the watermark extraction functionality. This is similar to API poisoning based defense for black-box (backdoor) watermarking \cite{Zhang2023Categorical}, but it is formulated herein for box-free watermarking. The proposed DGS is derived as follows.

\subsection{Gradient Reorientation}
\label{sec:gr}
For the ease of presentation, let $Z \triangleq \mathbb{D}[\mathbb{R}(Y)]$ (see Figure \ref{fig:gradient_based_attack}) and let $Z^{\ast}$ be the altered output that is returned to the user. According to (\ref{eq:r_loss}) and (\ref{eq:attack_gradient}), we have
\begin{equation}
\nabla {\mathcal{L}_{{\text{Removal}}}} \propto  \frac{{\partial {\mathcal{L}_{\text{Removal}}}}}{{\partial \mathbb{R}(Y)}} = 2{\left( Z - {W_0} \right)^T}\frac{\partial Z}{{\partial \mathbb{R}(Y)}},\label{eq:original_gradient}
\end{equation}
where $\{\cdot\}^T$ is the transpose operator. The training of $\mathbb{R}$ requires the curation of a set of $Y$ obtained from querying the encapsulated $\mathbb{M}$ and $\mathbb{E}$ using a set of $X_0$. During the initial training stage, $\mathbb{R}$ is not able to remove the watermark, yielding $Z \approx W$. To prevent $Z$ from eventually converging to $W_0$, we can reorient $Z$ into $Z^\ast$ when $Z \approx W$, so the true gradient direction can be protected. According to (\ref{eq:original_gradient}), the perturbed gradient component can be expressed as 
\begin{equation}
\left(\frac{\partial {\mathcal{L}_{\text{Removal}}}}{\partial \mathbb{R}(Y)}\right)^\ast = 2{\left( {Z^\ast - {W_0}} \right)^T}\frac{{\partial Z^\ast}}{{\partial \mathbb{R}(Y)}}, \label{eq:poison_gradient}
\end{equation}
which can be designed in such a way that the direction change is between $90$ and $180$ degree, i.e.,
\begin{align}
& {\left( {Z^\ast - {W_0}} \right)^T}\frac{{\partial {Z^\ast}}}{{\partial \mathbb{R}(Y)}} = - {\left( {Z - {W_0}} \right)^T}P\frac{{\partial Z}}{{\partial \mathbb{R}(Y)}} \notag\\
\Rightarrow & {\left( {Z^\ast - {W_0}} \right)^T}\frac{{\partial Z^\ast}}{{\partial Z}}= - {\left( {Z - {W_0}} \right)^T}P,\label{eq:z_inverse}
\end{align}
where the chain rule is applied to cancel $\partial Z / \partial \mathbb{R}(Y)$ and $P$ is a positive definite matrix. Note that if $P$ is the identity matrix $I$, then the above reorientation is simply gradient sign flipping ($180$ degree). Meanwhile, to preserve the output image quality, it is required that
\begin{equation}
Z^\ast \approx W. \label{eq:z_quality}
\end{equation}
Note that (\ref{eq:z_inverse}) is a first-order differential equation, and the solution for $Z^\ast$ has the following form  
\begin{equation}
Z^\ast = -P Z + C,\label{eq:z_form}
\end{equation}
where $C$ is independent of $Z$. To solve for $C$, substitute (\ref{eq:z_form}) into (\ref{eq:z_quality}), then we have
\begin{equation}
-P Z + C \approx W \Rightarrow C \approx (P+I)W, \label{eq:eq}
\end{equation}
and substituting (\ref{eq:eq}) into (\ref{eq:z_form}) yields
\begin{equation}
\label{eq:ultimate_trans}
Z^\ast = -PZ + (P+I)W,
\end{equation}
where the approximation is replaced by equality for implementation. The gradient reorientation in (\ref{eq:ultimate_trans}) is the essential component in the proposed DGS.

\subsection{DGS Protected Decoder API}
We now describe how the proposed gradient reorientation is incorporated in the API of decoder $\mathbb{D}$, as depicted in \mbox{Figure \ref{fig:gsp_architecture}}. Since $\mathbb{D}$ serves as a black-box watermark verifier to end users, we first discuss possible situations of its query denoted by $S$ without deploying DGS.

\textbf{Situation 1:} $S = Y = \mathbb{E}({\rm{Concat}}(\mathbb{M}(X_0), W)) \in \mathcal{Y}$. This means that $S$ is a processed and watermarked image from the black-box API of $\mathbb{M}$ and it corresponds to the benign query for watermark extraction. It then follows from Section \ref{sec:basic} that it returns $\mathbb{D}(S) = \mathbb{D}(Y) \approx W$. 

\textbf{Situation 2:} $S = \mathbb{R}(Y)$. This is the malicious query for gradient-based removal attack, but it is indistinguishable from \textbf{Situation 1} because (\ref{eq:a_fidelity_loss}) ensures that $\mathbb{R}(Y)$ is semantically identical to $Y$. However, since the true gradient has not been returned for $\mathbb{R}$ to learn watermark removal, the initial malicious query follows that $S=\mathbb{R}(Y)\in\mathcal{Y}$, and according to Section \ref{sec:gr}, it returns $\mathbb{D}(S) = \mathbb{D}[\mathbb{R}(Y)] = Z \approx W$.

\textbf{Situation 3:} $S \in \mathcal{Y}^\complement$. This corresponds to the benign query with a non-watermarked image, and it follows from Section \ref{sec:basic} that $\mathbb{D} (\mathcal{Y}^\complement) \approx W_0$ is returned.

According to the above situations and incorporating (\ref{eq:ultimate_trans}), we can summarize the response mechanism of DGS protected $\mathbb{D}$ API, denoted by $\mathbb{D}^\ast(S)$, as 
\begin{align}\label{eq:final_mechanism}
& \mathbb{D}^\ast(S) = \notag\\
& {\small{\left\{ {\begin{array}{*{20}{l}}
  { - P\mathbb{D}(S) + (P + I)W,} & {{\text{if }} {\rm{NC}}(\mathbb{D}(S), W) > 0.96}, \\ 
  {\mathbb{D}(S),}&{{\text{otherwise,}}} 
  \end{array}} \right.}}
\end{align}
where ${\rm{NC}(\cdot, \cdot)}$ is the normalized cross-correlation function, and we determine $\mathbb{D}(S) \approx W$ using the threshold of $0.96$ \cite{zhang2022deep}. The above mechanism means that for \textbf{Situations 1} and \textbf{2}, since they are indistinguishable, the API returns the gradient reoriented output, while for \textbf{Situation 3}, the API simply returns the original output. Note that the gradient reorientation in (\ref{eq:final_mechanism}) only adds imperceptible perturbations on the extracted watermark $W$, which does not affect the watermark verification, so its influence on normal users is negligible. However, such perturbation can effectively prevent the attacker from training $\mathbb{R}$ for watermark removal. 

We note that (\ref{eq:final_mechanism}) is the first solution to protecting the decoder $\mathbb{D}$ in box-free watermarking so as to prevent black-box model extraction. It is also a neat closed-form solution compared to gradient-based defense in other contexts, e.g., the recursive methods called prediction poisoning \cite{pp_2020_iclr_defense_model_stealing} and gradient redirection \cite{Mantas2022How}, for protecting deep classification models.

\subsection{Choice of $P$}
\label{sec:p_choice}
The positive definite matrix $P$ introduced in (\ref{eq:z_inverse}) is an essential component in the proposed reorientation process. The risk of omitting $P$, or equivalently setting $P=I$, is that the attacker can simply flip the gradient sign back to obtain the true gradient. In the situation where $P \neq I$, let the eigendecomposition be $P = Q^T \Lambda Q$, where $\Lambda$ is the diagonal eigenvalue matrix and $Q$ the eigenvector orthonormal matrix, then for a vector multiplied at its right-hand side, it first rotates the vector via $Q$, followed by scaling the vector elements by the eigenvalues in $\Lambda$ and finally reverting the rotation via $Q^T$. The rotation imposed by $Q$ is compensated in this process, but the rotation indirectly caused by scaling using $\Lambda$ is not. For example, as just one use case, we consider $Q=I$ and thus $P=\Lambda$ is a diagonal matrix with all-positive elements. As long as these elements are unequal, $-P$ in (\ref{eq:z_inverse})  can incur a $90$ to $180$ degree rotation. Note that the attacker can still flip the gradient sign, while this will result in a gradient deviated by $0$ to $90$ degree. To further prevent the learning of $\mathbb{R}$, we can reduce the gradient norm by setting $0 < \Lambda_i \ll 1$, where $\Lambda_i$ is the $i$th diagonal element, so that even the attacker succeeds in recovering the true gradient direction, the rate of learning becomes negligibly small.

\begin{figure*}[!ht]
    \centering
    \begin{subfigure}{0.33\textwidth}
        \centering
        \includegraphics[width=\linewidth]{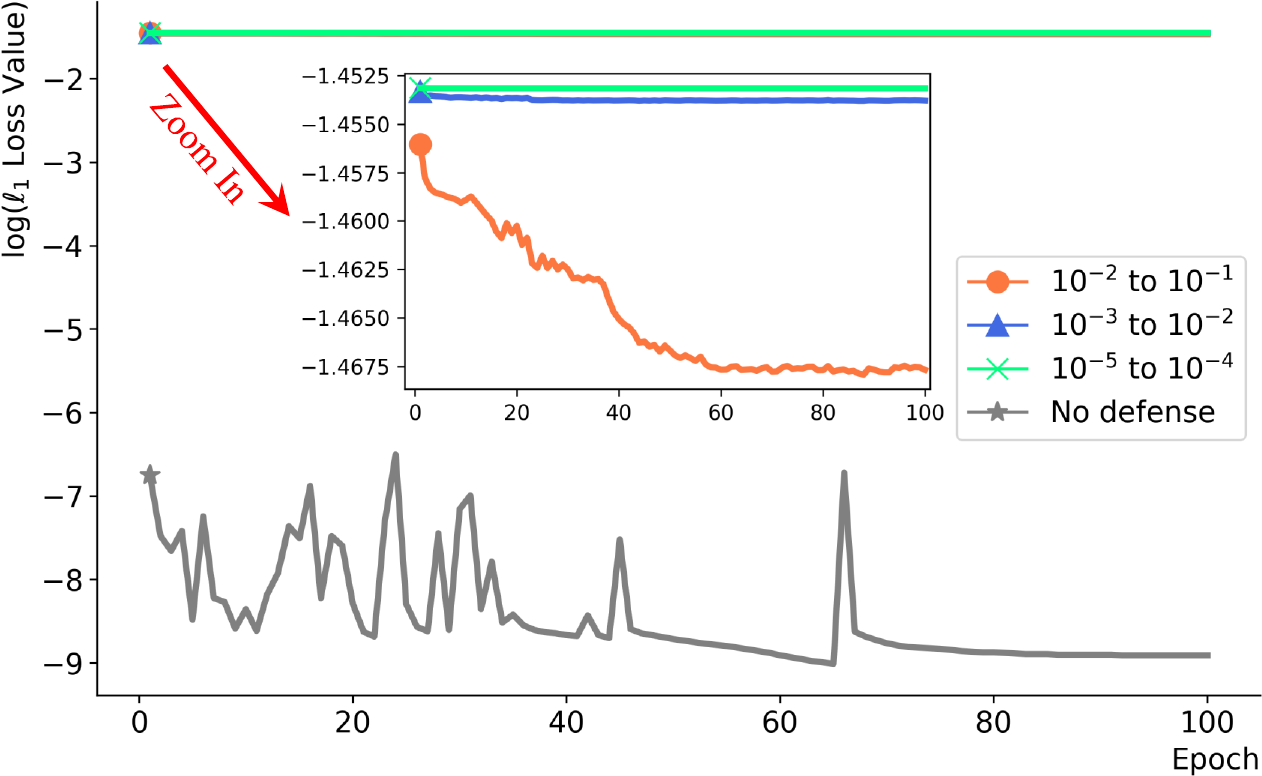}
        \caption{$\ell_1$ loss}
    \end{subfigure}
    \hfill
    \begin{subfigure}{0.33\textwidth}
        \centering
        \includegraphics[width=\linewidth]{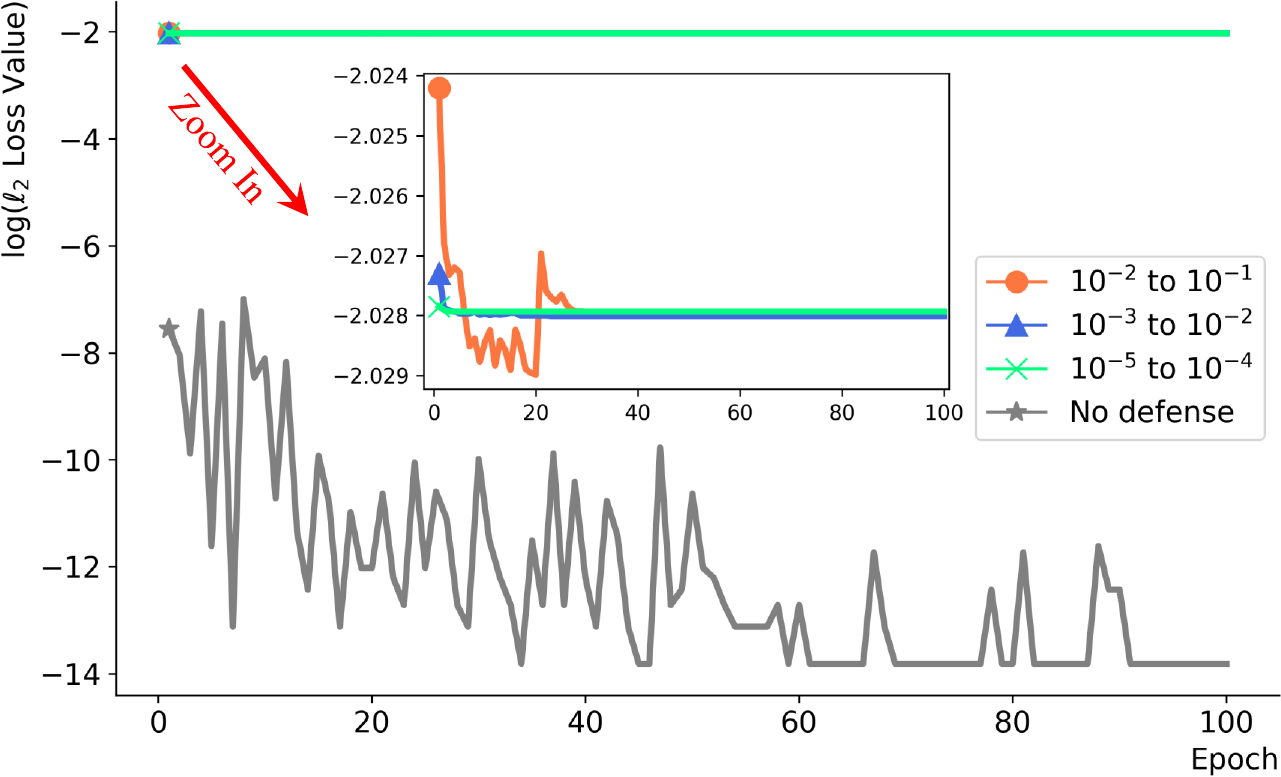}
        \caption{$\ell_2$ loss}
    \end{subfigure}
    \hfill
    \begin{subfigure}{0.33\textwidth}
        \centering
        \includegraphics[width=\linewidth]{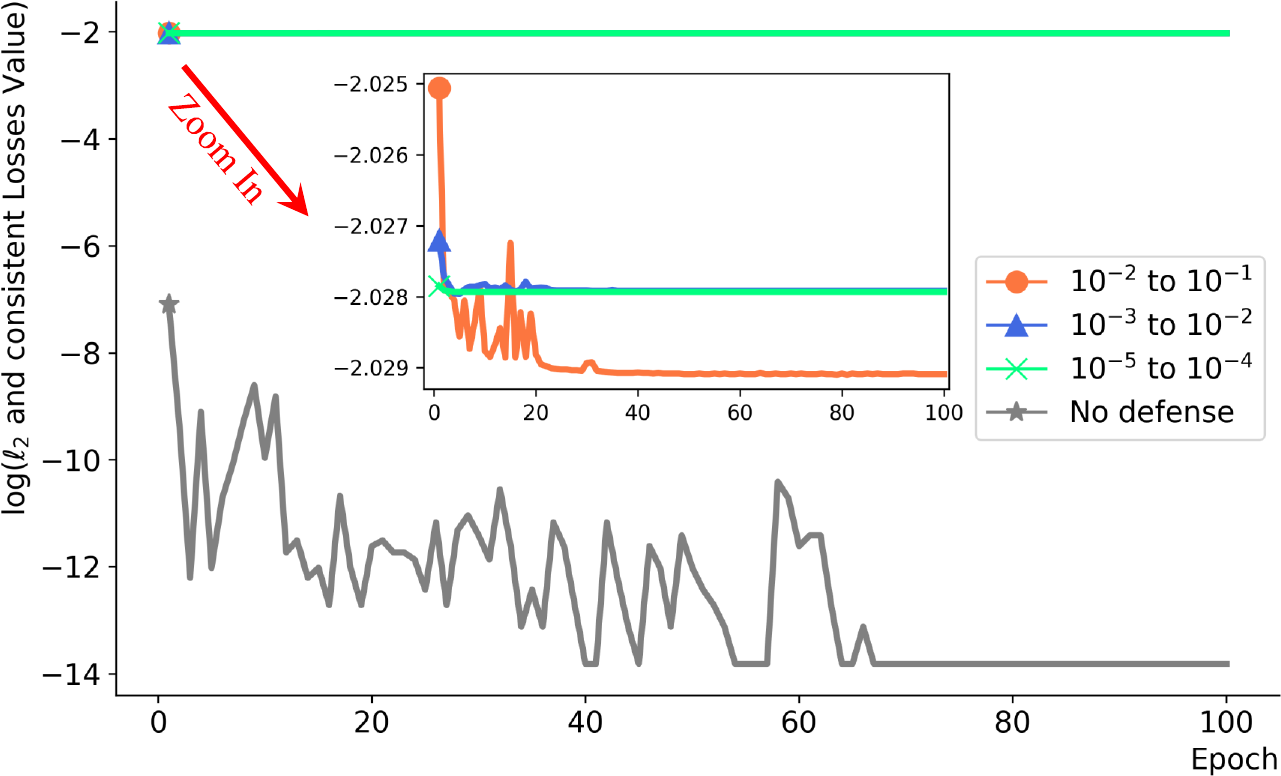}
        \caption{$\ell_2$ and consistent losses}
    \end{subfigure}
    
    \begin{subfigure}{0.33\textwidth}
        \centering
        \includegraphics[width=\linewidth]{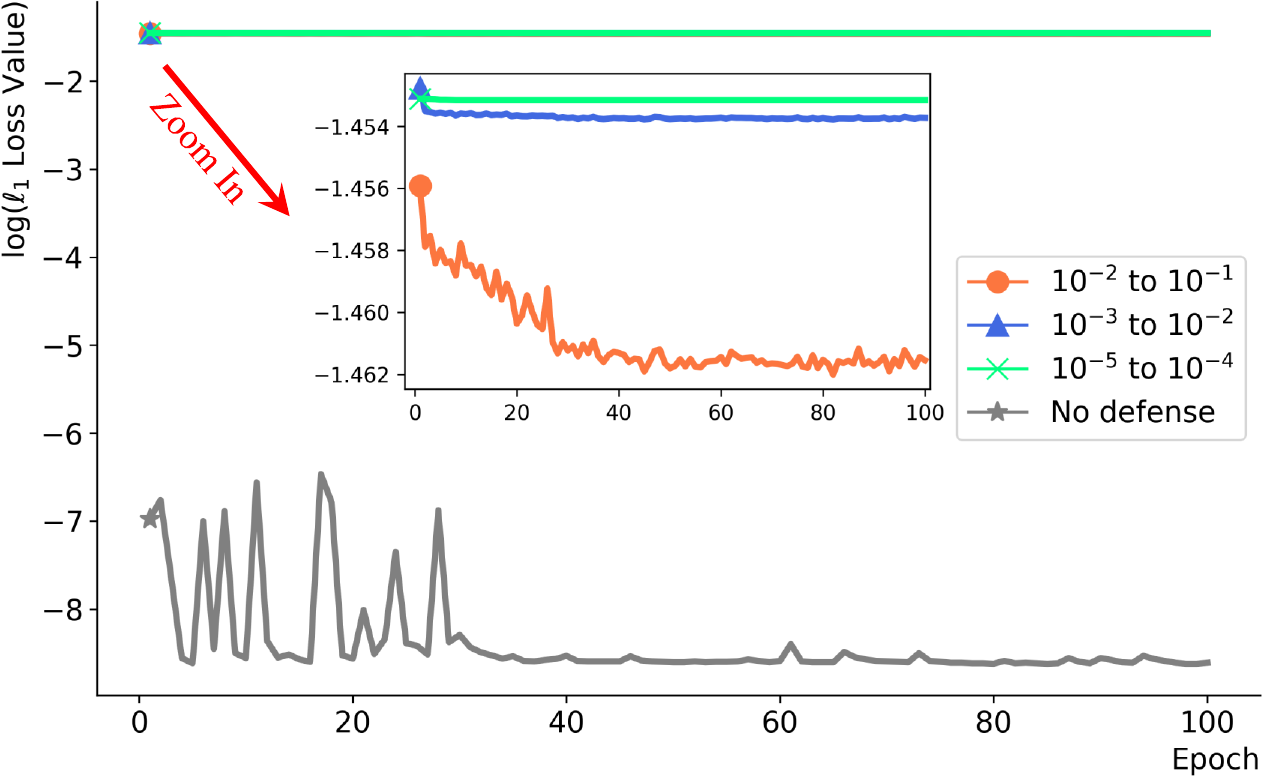}
        \caption{$\ell_1$ loss}
    \end{subfigure}
    \hfill
    \begin{subfigure}{0.33\textwidth}
        \centering
        \includegraphics[width=\linewidth]{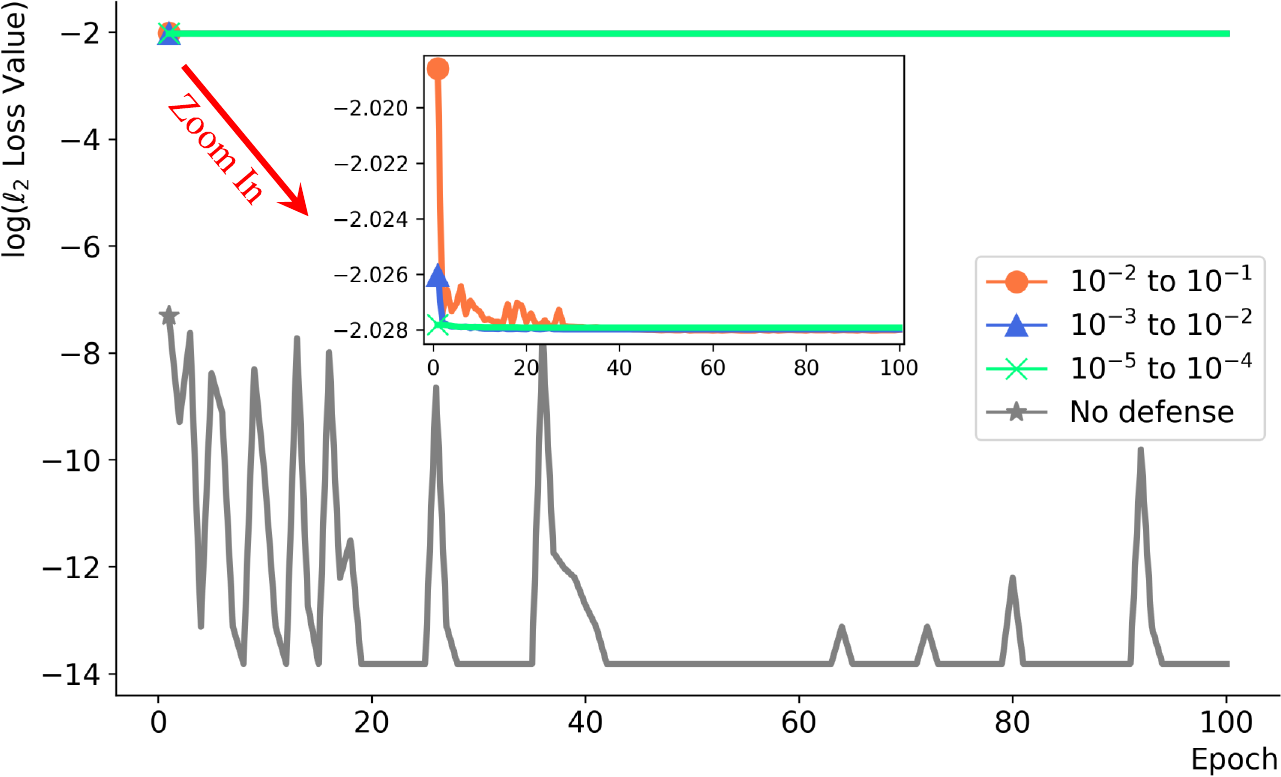}
        \caption{$\ell_2$ loss}
    \end{subfigure}
    \hfill
    \begin{subfigure}{0.33\textwidth}
        \centering
        \includegraphics[width=\linewidth]{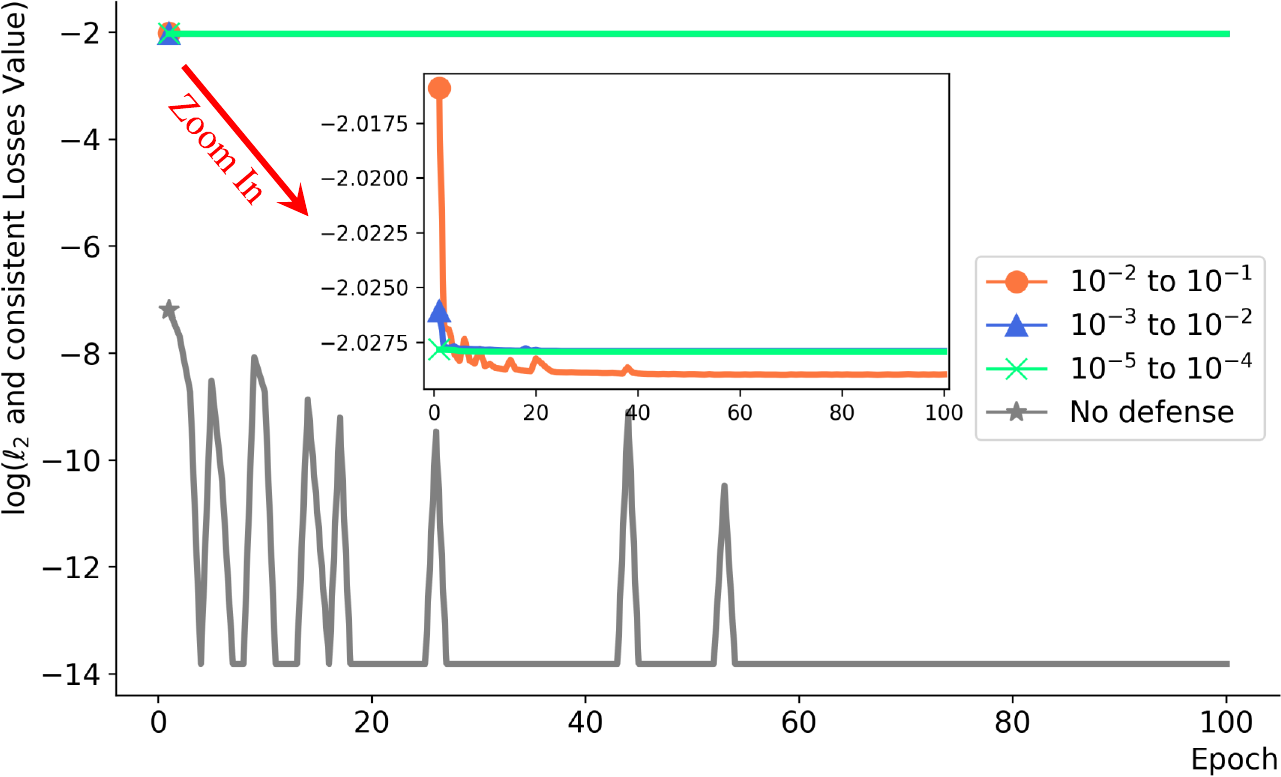}
        \caption{$\ell_2$ and consistent loss}
    \end{subfigure}
    
    \caption{Demonstration of the convergence behavior of attacker's removal loss functions when training $\mathbb{R}$, under different choices of $P$. The $1$st row is deraining and the $2$nd row is style transfer. The $\ell_1$ loss is $\|Z^{\ast} - W_0\|_1^2$, the $\ell_2$ loss is $\|Z^{\ast} - W_0\|_2^2$, while the consistent loss is incorporated from \cite{zhang2020model}. The loss corresponding to no defense is $\|Z - W_0\|_2^2$.}
    \label{fig:convergence_various_losses}
\end{figure*}

\section{Experimental Results}
\label{sec:experiment}
In this section, we present the experimental results to verify the effectiveness of our proposed DGS. We select the state-of-the-art box-free watermarking model \cite{zhang2024robust} to implement our defense. Notably, \cite{zhang2024robust} is an extended version of \cite{zhang2022deep}, addressing the vulnerabilities of their watermarking scheme to image augmentation attacks, while both share the same watermark encoder and decoder. 

\subsection{Experimental Setting}
\subsubsection{Datasets}
We consider two representative image-to-image tasks, i.e., image deraining (classic low-level image processing) and style transfer (a high-level computer vision task). For both tasks, we use the PASCAL VOC dataset \cite{everingham2010pascal} with different data splits. For image deraining, the data corresponds to $X$ in Figure \ref{fig:victim_architecture}. We uniformly split the $12,000$ training images into two equal parts, each containing $6,000$ for victim model training and remover training, respectively. Rainy images corresponding to $X_0$ in Figure \ref{fig:victim_architecture} are generated using the algorithm in \cite{zhang2018density}. For style transfer, the PASCAL VOC data are treated as $X_0$ and similarly divided for victim and remover training. The style transfer algorithm in \cite{li2018learning} is employed to generate $X$. Additionally, to reduce the computational complexity, we resize all images to $256 \times 256$ grayscale.

\subsubsection{Metric}
For fidelity evaluation, we use peak signal-to-noise ratio (PSNR) and multi-scale structural similarity index (MS-SSIM) \cite{wang2003multiscale} to measure the similarity between two images. For robustness evaluation, we use the success rate of defense, denoted by $\text{SR}$, which is the ratio of the number of images with embedded watermarks successfully extracted over the total number of watermarked images under going through removal attacks. 


\subsubsection{Implementation Detail}
We follow the model architecture, hyperparameters, and training process of victim model in \cite{zhang2024robust}, including $\mathbb{M}$, $\mathbb{E}$, and $\mathbb{D}$. The gradient-based remover $\mathbb{R}$ is implemented using a UNet architecture \cite{ronneberger2015u}. Both models are trained from scratch for $100$ epochs using the Adam optimizer with a learning rate of $0.0002$. Weighting parameters $\alpha_1$, $\alpha_2$, $\beta_1$, and $\beta_2$ in (\ref{eq:loss_joint}) and (\ref{eq:loss_attacker}) are set equally to $1$.

\subsection{Convergence of Removal Loss}
\label{sec:convergence_r_loss}
We first verify the effectiveness of the proposed DGS in terms of preventing $\mathbb{R}$ from learning to remove the watermark, reflected by the loss behavior during the training of $\mathbb{R}$. Due to DGS, the actual removal loss obtained by the attacker is modified from (\ref{eq:r_loss}) to 
\begin{equation}
    \mathcal{L}^{\text{A}\ast}_\text{Removal}  =  \|Z^{\ast} - W_0\|_2^2,
\end{equation}
when the $\ell_2$-norm is used. It is worth noting that the proposed DGS is derived based on the $\ell_2$-norm removal loss function. In fact, the attacker can use other removal loss functions, such as the $\ell_1$-norm and $\ell_2$-norm plus consistent loss \cite{zhang2020model}. This reflects the real-world situation of removal loss mismatch, which is also considered in our experiments, while the fidelity loss is consistently the $\ell_2$-norm in (\ref{eq:a_fidelity_loss}). The results for deraining and style transfer tasks are presented in the first and second rows, respectively, in Figure \ref{fig:convergence_various_losses}.

\begin{figure}[!t]
    \centering
    \includegraphics[width=.85\linewidth]{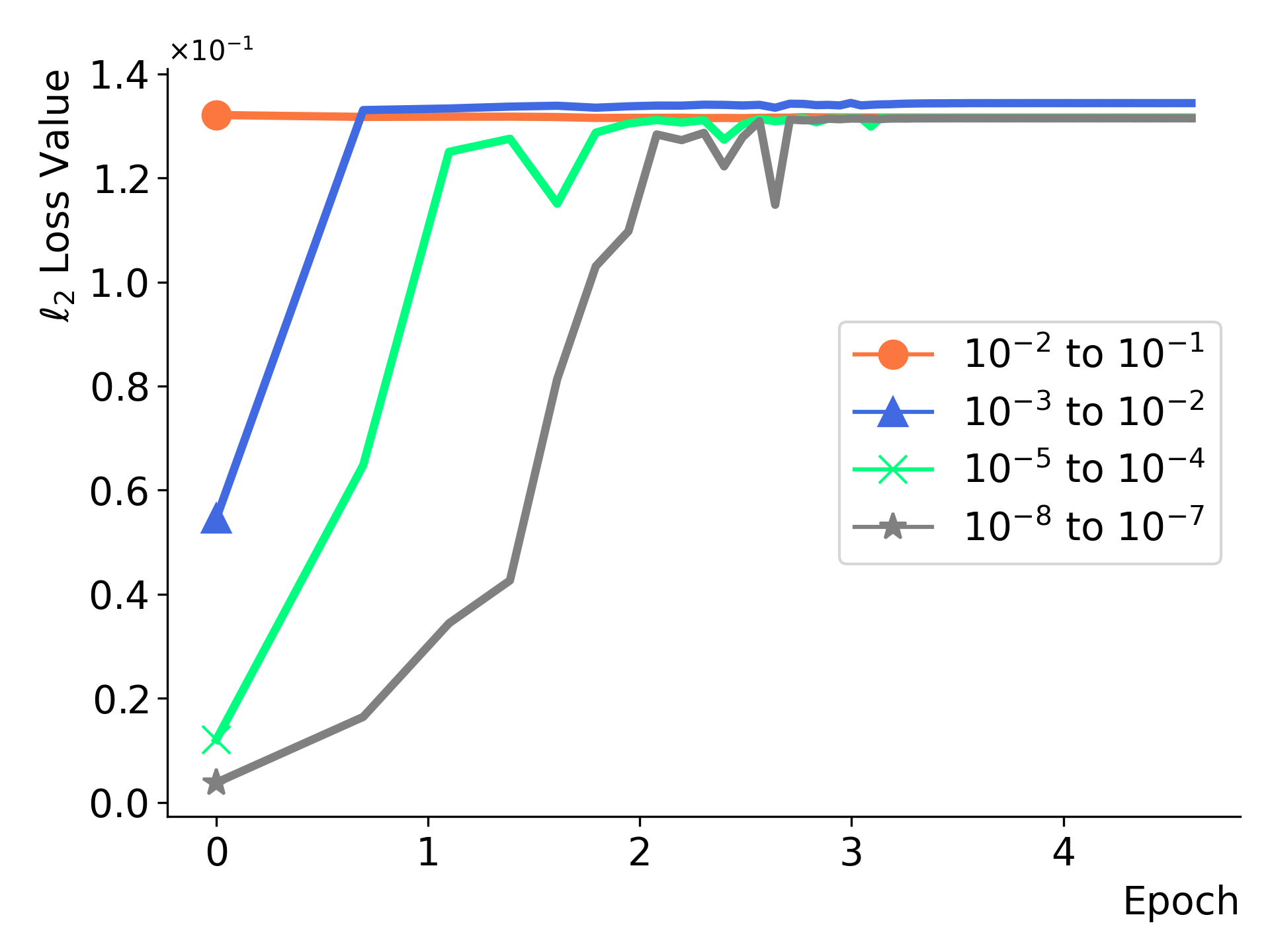}
    \vspace*{-6pt}
    \caption{Demonstration of the convergence behavior of the true loss function $\|Z - W_0\|_2^2$ after deploying the proposed DGS, under different choices of $P$, and deraining is considered as an example.}
    \label{fig:attacker_flip}
\end{figure}

For both tasks, when no defense is deployed, all loss values converge to the zero level, $10^{-8}$ and $10^{-14}$, respectively, which verifies the effectiveness of the gradient-based removal attack. It is also observed that the use of the consistent loss leads to a smoother convergence curve. In contrast, with DGS deployed, none of the loss functions can be reduced. It can be seen from the zoom-in versions that all the loss values show a decreasing trend but in a negligibly small range, indicating the successful prevention of $\mathbb{R}$ from learning to remove the watermark. The results also verify that the proposed defense based on the $\ell_2$-norm removal loss can be well generalized against other loss functions.

To further verify the ineffectiveness of the training of $\mathbb{R}$, we present the loss between the raw output of $\mathbb{D}$, i.e., $Z$, and $W_0$ when $\mathbb{R}$ is trained in presence of DGS, and the results are shown in Figure \ref{fig:attacker_flip}, where deraining is considered as an example. It can be seen that for all choices of $P$, the loss values increase and converge to a high level corresponding to failure of watermark removal. 

\subsection{Effectiveness}
\label{sec:effectiveness}
\begin{figure}[!t]
    \centering
    \includegraphics[width=1.0\linewidth]{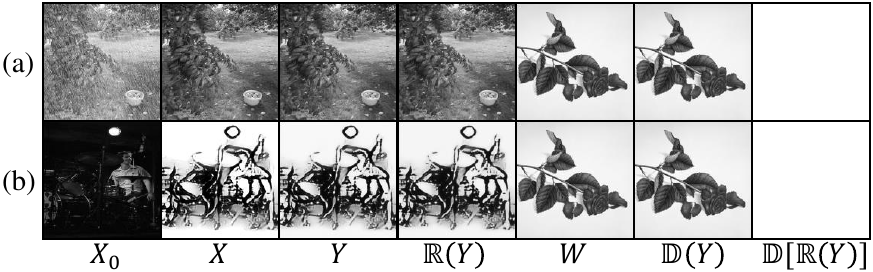}
    \caption{Demonstration of the gradient-based watermark removal attack without defense. (a) Deraining. (b) Style transfer.}
    \label{fig:effectiveness_no_defense}
\end{figure}

\begin{figure}[!t]
    \centering
    \includegraphics[width=1.0\linewidth]{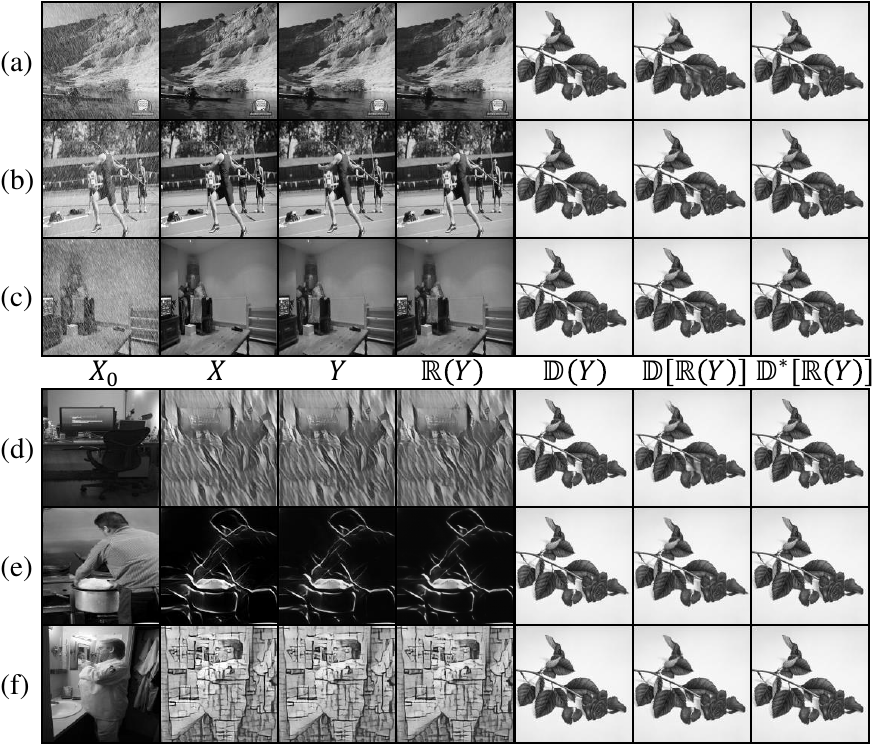}
    \caption{Demonstration of the gradient-based watermark removal attack with the proposed DGS defense, where (a)--(c) are deraining, (d)--(f) are style transfer, and $W$ is the same as in Figures \ref{fig:effectiveness_no_defense}. The removal loss functions used by the attacker are $\ell_1$ loss in (a) and (d), $\ell_2$ loss in (b) and (e), and $\ell_2$ plus consistent in (c) and (f).}
    \label{fig:effectiveness}
\end{figure}

Illustrative image and watermark examples for both deraining and style transfer tasks before and after the proposed DGS are presented in Figures \ref{fig:effectiveness_no_defense} and \ref{fig:effectiveness}, respectively. The images from left to right are the to-be-processed $X_0$, processed non-watermarked $X$, processed and watermarked $Y$, watermarked attacked by remover $\mathbb{R}(Y)$, original watermark $W$, decoded watermark without attack $\mathbb{D}(Y)$, decoded watermark after attack $\mathbb{D}[\mathbb{R}](Y)$ (or equivalently $Z$), and DGS perturbed result  $\mathbb{D}^\ast[\mathbb{R}](Y)$ (or equivalently $Z^\ast$). It can be observed that without defense, the gradient-based attack successfully removes the watermark from $Y$ in both tasks, leading to nearly all-white images at the output of $\mathbb{D}$. In contrast, with the proposed DGS, the attacker cannot remove the watermark, not only with matched loss (Figure \ref{fig:effectiveness} (b) and (e)) but also with mismatched losses (Figure \ref{fig:effectiveness} other subfigures). Additionally, comparing the last two columns, DGS preserves decoder output image quality with imperceptible difference between $\mathbb{D}[\mathbb{R}(Y)]$ and $\mathbb{D}^\ast[\mathbb{R}(Y)]$, allowing legitimate queries while preventing the training of $\mathbb{R}$.


\subsection{Robustness}
\label{sec:robustness}
Despite that the perturbed output $\mathbb{D}^\ast[\mathbb{R}(Y)]$ is returned, the attacker has the freedom to further process it before using it to update $\mathbb{R}$ parameters with the hope of rendering potential defenses ineffective. Here, we consider three types of further processing imposed by an attacker:
\begin{itemize}
\item JPEG compression and noise addition.
\item Lattice attack \cite{liu2023erase}.
\item Gradient sign flipping. 
\end{itemize}
Among them, JPEG compression and noise addition (using i.i.d. white Gaussian noise) are intuitive image quality degradation operations, the lattice attack replaces pixels by random values according to a fixed step for watermark removal, while gradient sign flipping can be launched if the attacker observes no reduction of the removal loss and thus believe that the gradient has been flipped. 

Results for the first two attacks are presented in \mbox{Tables \ref{tab:JPEG_compression}--\ref{tab:lattice_attack}}, where PSNR and MS-SSIM are used to measure the similarity between the API returned gradient-reoriented output $\mathbb{D}^\ast[\mathbb{R}(Y)]$ and the attacked version. It can be seen that even with a JPEG compression factor of $10\%$ or a lattice attack that randomly alters every $1$ out of $3$ pixels (step is $2$), the success rate of watermark extraction is still $100\%$. Guaranteed success is preserved with $10$ dB noise addition, while the rate reduces to $58\%$ under $0$ dB noise for style transfer only, and this is with substantial image quality degradation. Generally, the proposed DGS demonstrates promising robustness against both normal and advanced attacks.

\begin{table}[!t]
  \centering
  \caption{Robustness test of DGS against JPEG compression, where $10^{-5}<\Lambda_i<10^{-4}$, PSNR is in dB, and $0\leq\text{MS-SSIM},\text{SR}\leq 1$.}
  \setlength{\tabcolsep}{1pt}
  \label{tab:JPEG_compression}
  \begin{tabular}{c|ccc|ccc}
  \hline
  \hline
  \multirow{2}{*}{Factor} & \multicolumn{3}{c|}{Deraining} & \multicolumn{3}{c}{Style Transfer} \\
  \cline{2-7}
  & PSNR$\uparrow$ & MS-SSIM$\uparrow$ & SR$\uparrow$ & PSNR$\uparrow$ & MS-SSIM$\uparrow$ & SR$\uparrow$  \\
  \hline
  $10\%$ & $28.5748$ & $0.9672$ & $1.00$ & $28.5776$ & $0.9673$ & $1.00$ \\ 
  $20\%$ & $30.8380$ & $0.9855$ & $1.00$ & $30.8364$ & $0.9855$ & $1.00$ \\ 
  $30\%$ & $32.0850$ & $0.9902$ & $1.00$ & $32.0845$ & $0.9902$ & $1.00$ \\ 
  $40\%$ & $33.0321$ & $0.9927$ & $1.00$ & $33.0278$ & $0.9927$ & $1.00$ \\ 
  \hline
  \hline
  \end{tabular}
\end{table}

\begin{table}[!t]
  \centering
  \caption{Robustness test of DGS against WGN addition, where $10^{-5}<\Lambda_i<10^{-4}$, noise level and PSNR are in dB, and $0\leq\text{MS-SSIM},\text{SR}\leq 1$.}
  \setlength{\tabcolsep}{1pt}
  \label{tab:Gaussian_noise}
  \begin{tabular}{c|ccc|ccc}
  \hline
  \hline
  \multirow{2}{*}{\begin{tabular}{@{}c@{}} Noise \\ Level \end{tabular}} & \multicolumn{3}{c|}{Deraining} & \multicolumn{3}{c}{Style Transfer} \\
  \cline{2-7}
  & PSNR$\uparrow$ & MS-SSIM$\uparrow$ & SR$\uparrow$ & PSNR$\uparrow$ & MS-SSIM$\uparrow$ & SR$\uparrow$  \\
  \hline
  $0$ & $1.7791$ & $0.3042$ & $1.00$ & $1.7768$ & $0.3041$ & $0.58$ \\ 
  $10$ & $11.7799$ & $0.5613$ & $1.00$ & $11.7809$ & $0.5612$ & $1.00$ \\ 
  $20$ & $21.7774$ & $0.8022$ & $1.00$ & $21.7801$ & $0.8022$ & $1.00$ \\ 
  $30$ & $31.7781$ & $0.9573$ & $1.00$ & $31.7784$ & $0.9573$ & $1.00$ \\ 
  \hline
  \hline
  \end{tabular}
  \end{table}

\begin{table}[!t]
  \centering
  \caption{Robustness test of DGS against lattice attack \cite{liu2023erase}, where $10^{-5}<\Lambda_i<10^{-4}$, PSNR is in dB, and $0< \text{MS-SSIM},\text{SR}<1$.}
  \setlength{\tabcolsep}{1pt}
  \label{tab:lattice_attack}
  \begin{tabular}{c|ccc|ccc}
  \hline
  \hline
  \multirow{2}{*}{Step} & \multicolumn{3}{c|}{Deraining} & \multicolumn{3}{c}{Style Transfer} \\
  \cline{2-7}
  & PSNR$\uparrow$ & MS-SSIM$\uparrow$ & SR$\uparrow$ & PSNR$\uparrow$ & MS-SSIM$\uparrow$ & SR$\uparrow$  \\
  \hline
  $2$ & $12.3766$ & $0.6391$ & $1.00$ & $12.3766$ & $0.6391$ & $1.00$ \\ 
  $6$ & $21.9368$ & $0.8275$ & $1.00$ & $21.9370$ & $0.8275$ & $1.00$ \\ 
  $11$ & $26.8371$ & $0.9263$ & $1.00$ & $26.8373$ & $0.9263$ & $1.00$ \\ 
  $16$ & $30.6777$ & $0.9655$ & $1.00$ & $30.6777$ & $0.9655$ & $1.00$ \\ 
  \hline
  \hline
  \end{tabular}
  \end{table}

To provide more insights, let $F$ be a generic additive interference resulted from the attacker's further operation, and $F$ is independent of $Z$, then
(\ref{eq:ultimate_trans}) is modified to
\begin{equation}
    \label{eq:interfered_shield}
    Z^{\ast}_{\text{Interf}} = -PZ + (P+I)W + F,
\end{equation}
and the attacker's gradient component is modified from (\ref{eq:poison_gradient}) to
\begin{align}
    & 2{\left( {Z^{\ast}_{\text{Interf}} - {W_0}} \right)^T}\frac{{\partial Z^\ast_{\text{Interf}}}}{{\partial \mathbb{R}(Y)}} \notag\\
    & = -2{\left[ {Z - \left({W_0} - F \right)} \right]^T}P\frac{{\partial Z}}{{\partial \mathbb{R}(Y)}}.
\end{align}
This means that with the deployment of DGS, the additive interference forces the remover output to diverge from $W_0 - F$ instead of diverging from the original $W_0$, which still cannot undermine the defense. Additionally, similar performance between deraining and style transfer tasks indicates that DGS is insensitive to data distribution and can be generalized to other image-to-image tasks. 

\begin{figure}[!t]
    \centering
    \includegraphics[width=.9\linewidth]{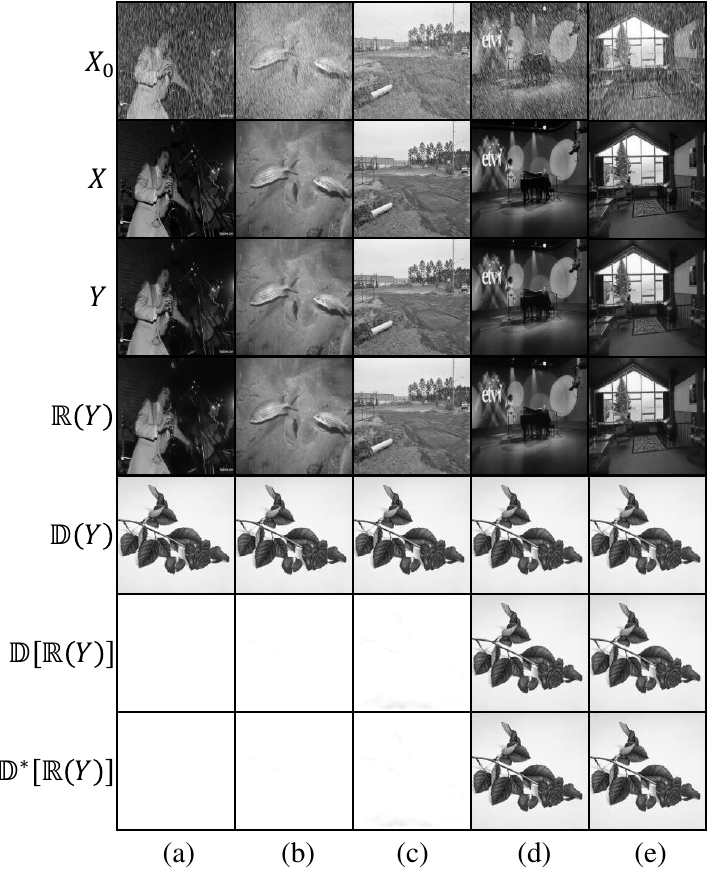}
    \caption{Demonstration of the robustness of the proposed DGS when the attacker applies gradient sign flipping, where $W$ is the same as in Figures \ref{fig:effectiveness_no_defense}, deraining is used as a example, and $\Lambda_i$ is randomly sampled within (a) $[10^{-4}, 10^{-3}]$, (b) $[10^{-5}, 10^{-4}]$, (c) $[10^{-6}, 10^{-5}]$, (d) $[10^{-7}, 10^{-6}]$, and (e) $[10^{-8}, 10^{-7}]$.}
    \label{fig:robustness_attacker_flip}
\end{figure}

For gradient sign flipping, the reorientation can be partially compensated, not fully, because it is not strictly $90$ degree. Due to this, the attacker-flipped gradient will contain the true gradient component and enable $\mathbb{R}$ to learn watermark removal, though not most efficiently. However, the proposed small values of $\Lambda_i$ can effectively reduce the learning rate. This is illustrated in Figure \ref{fig:robustness_attacker_flip}. It can be seen that when $\Lambda_i$ is within the interval $[10^{-7},10^{-6}]$ or smaller, DGS remains robust to gradient-based removal attacks.

\subsection{Limitation}
\label{sec:limitation}
We have discussed and experimented with attacker's intuitive and practical countermeasure of flipping the gradient sign before updating $\mathbb{R}$ parameters, if it is observed that the removal loss cannot be reduced. It is shown that the proposed DGS can still withstand such gradient flipping thanks to the small diagonal values of $P$, while here we provide a further discussion about the the potential weakness of DGS.

According to (\ref{eq:ultimate_trans}), the attacker can fully overcome DGS if the hidden $Z$ can be recovered from the observed $Z^\ast$. To achieve so, the inverse of (\ref{eq:ultimate_trans}) is given by
\begin{equation}
    Z = - P^{-1}Z^{\ast} + \left(I + P^{-1}\right)W, \label{eq:limitation}
\end{equation}
requiring the knowledge of $W$ and $P$. While $W$ can be estimated by querying $\mathbb{D}$ using a watermarked image $Y \in \mathcal{Y}$, it is difficult to guess $P$. However, the attacker may simply set $P=I$ and replace $W$ by $\mathbb{D}^{\ast}(Y)$ in (\ref{eq:limitation}), which yields an estimate of $Z$ given by $\Tilde{Z} = - Z^{\ast} + 2 \mathbb{D}^{\ast}(Y)$. According to our experiments, such an approximation suffers from performance degradation, but it remains open for the attacker to develop more advanced attacks to improve the estimation of $Z$. We note that the existing gradient-based defense methods, e.g., \cite{pp_2020_iclr_defense_model_stealing} and  \cite{Mantas2022How}, are also vulnerable to gradient sign flipping, indicating that incurring a gradient rotation between $90$ to $180$ degree may not be sufficient for protection, which calls for further investigation.

\section{Conclusion}
\label{sec:conclusion}
Existing box-free watermarking methods for image-to-image models use a dedicated decoder $\mathbb{D}$ for watermark extraction directly from watermarked images. Since $\mathbb{D}$ is coupled with the protected watermark encoder $\mathbb{E}$, its black-box querying process makes the watermarking mechanism vulnerable and thus be exploited for watermark removal. Motivated by this observation, we have developed a gradient-based removal attack which can remove state-of-the-art box-free watermarks. To address this vulnerability, we have then proposed the decoder gradient shield (DGS) framework in the black-box API of $\mathbb{D}$ and derived a closed-form solution. DGS can effectively reorient and rescale the gradient of watermarked queries through a newly introduced positive definite matrix $P$. With proper choices of the eigenvalues of $P$, it is shown that DGS can prevent $\mathbb{R}$ from learning to remove watermarks. We have conducted extensive experiments and verified the effectiveness of our proposed DGS on both image deraining and style transfer tasks. Further investigation on improving the robustness of DGS in terms of preventing reverse engineering of the true gradient is still needed.

{
    \small
    \bibliographystyle{ieeenat_fullname}
    \bibliography{main}
}


\end{document}